\documentclass[sigconf]{acmart}
\usepackage[font=small,labelfont=bf,skip=0pt]{caption}
\usepackage{enumerate}
\usepackage{amsmath}
\usepackage{amsfonts}
\usepackage{graphicx}
\usepackage{amsmath}
\usepackage{lipsum}
\usepackage{multirow}
\usepackage{tabularx}
\usepackage{mathtools}
\usepackage{balance}
\usepackage{subfigure}
\usepackage{bbm}
\usepackage{color}
\usepackage{float}
\usepackage{listings}
\usepackage{color}
\usepackage{booktabs}
\usepackage{tablefootnote}
\usepackage{url}
\usepackage{threeparttable}
\usepackage{pifont}
\usepackage{tcolorbox}

\definecolor{ao(english)}{rgb}{0.0, 0.5, 0.0}

\newcommand\possiblebreak{\ifhmode\unskip\space\hfil\penalty0\hfilneg\fi}

\usepackage{adjustbox}
\renewcommand{\paragraph}[1]{\textbf{#1} }

\usepackage{algorithmic}
\usepackage[ruled,vlined]{algorithm2e}
\SetAlFnt{\small}

\setcopyright{none}
\renewcommand\footnotetextcopyrightpermission[1]{} 
\begin{document}
\sloppy
\title{Autonomy 2.0: The Quest for Economies of Scale}

\author{Shuang~Wu, Bo~Yu, Shaoshan~Liu, Yuhao~Zhu}


\maketitle

\section{Introduction}
\label{sec:intro}

With the advancement of robotics and AI technologies in the past decade, we have now entered the age of autonomous machines. In this new age of information technology, autonomous machines, such as service robots, autonomous drones, delivery robots, and autonomous vehicles, rather than humans, will provide services ~\cite{liu2022rise}. The rise of autonomous machines promises to completely transform our economy. However, after more than a decade of intense R\&D investments, autonomy has yet to deliver its promise~\cite{gates2007robot}. 

In this article, through examining the technical challenges and economic impact of the digital economy, we argue that scalability is both highly necessary from a technical perspective and significantly advantageous from an economic perspective, thus is the key for the autonomy industry to achieve its full potential. Nonetheless, the current development paradigm, dubbed Autonomy 1.0, scales with the number of engineers, instead of with the amount of data or compute resources, hence preventing the autonomy industry to fully benefit from the economies of scale, especially the exponentially cheapening compute cost and the explosion of available data. We further analyze the key scalability blockers and explain how a new development paradigm, dubbed Autonomy 2.0, can address these problems to greatly boost the autonomy industry.

\section{Scalability of the Digital Economy}
\label{sec:scale}

The digital economy refers to the use of information technology to create, market, distribute, and consume goods and services. It has been the key driving force for the world's economic growth in the past two decades.
Consider the internet industry, for instance.
The internet industry has accounted for 21\% of the GDP growth in mature economies from 2005 to 2010 ~\cite{manyika2011great}.
In 2019, the internet industry contributed \$2.1 trillion to the U.S. economy, about 10\% of the U.S. GDP, and is the fourth largest industry of the U.S. economy (behind only real estate, government, and manufacturing) ~\cite{hooton2019measuring}.
Along with its contribution to economy, the internet industry provides nearly 6 million direct jobs, accounting for 4\% of U.S. employments.

Two key forces fuel the continuous growth of the digital economy, both of which have to do with \textit{scalability}:

\begin{itemize}
\item The commoditization of computing power, as exemplified by Moore's law ~\cite{schaller1997moore}, is the greatest driving force behind the digital industry.
The most successful digital economy companies have developed core technology stacks that are scale by the available compute resources and data, not by the size of their engineering teams. One remarkable example is WhatsApp: when acquired by Facebook for \$19 billion, WhatsApp had only 32 engineers serving over 450 million users.

\item The breakthrough of artificial intelligence in the last decade has demonstrated that, in addition to many technical improvements and tuning, scaling neural network models and training datasets has been our most effective strategy for achieving continuous performance gains ~\cite{sutton2019bitter}.
\end{itemize}

Autonomy technologies such as those found in autonomous driving are widely seen as the pillar of the next digital economy era.
However, today's autonomous machines technologies, dubbed Autonomy 1.0, represent everything a scalable industry should \textit{not} do.

\begin{figure}
\centering
\includegraphics[width=0.9\columnwidth]{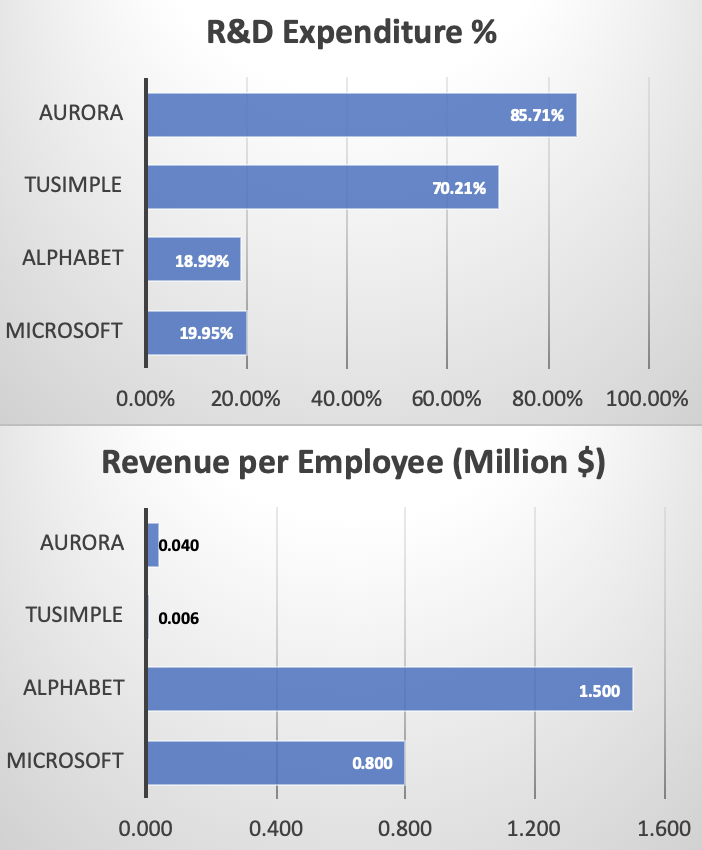}
\caption{Revenue and Expenditure Comparison: Digital Economy Companies vs. Autonomy Companies}
\vspace{10pt}
\label{fig:exprev}
\end{figure}

To illustrate the problem facing autonomous driving companies, Figure \ref{fig:exprev} analyzes the R\&D expenditures and revenue per employee of two leading public digital economy companies, Microsoft representing the software industry and Alphabet representing the internet industry, and two public autonomous driving companies, TuSimple representing the robot truck industry and Aurora representing the robotaxi industry. We selected these autonomous driving companies for the accessibility of their financial data. 

Both Alphabet and Microsoft spend less than 20\% of their total operating expenditures on R\&D. For instance, Google employs less than 30,000 engineers while serving over 4.3 billions of users. Their scalability is mainly constrained by available compute resources and data instead of by the number of engineers.
In comparison, both TuSimple and Aurora spend more than 70\% of their operating expenditures on R\&D. Often, to reach new users or to deploy services to new locations, autonomous driving companies need to pour additional R\&D resources to re-calibrate their existing technology stacks to adapt to new environments.
Hence, their scalability is constrained by R\&D investment or, more directly, the number of engineers. 
As a result, Alphabet and Microsoft are able to generate \$1.5 million and \$0.8 million of revenue per employee respectively while maintaining a high growth rate, whereas TuSimple and Aurora generate negligible revenue per employee and struggle with growth. For the autonomy industry to achieve economies of scale, we have to revolutionize the R\&D paradigm.

In following sections, we will describe key scalability issues with Autonomy 1.0, and outline promising solutions that are already at the horizon to achieve scalability in Autonomy 2.0.

\section{Autonomy 1.0: The End of The Road of An Aging architecture}
\label{sec:a1}

Current commercial autonomous driving systems mostly inherited the software architecture from competitors in the DARPA Grand Challenges between 2005 and 2007 ~\cite{thrun2006stanley}. This software architecture, while represented a great leap of autonomy technology at the time, has showed its age and become difficult to scale after more than a decade of intense industry efforts to improve and adapt. 

\begin{figure}
\centering
\includegraphics[width=\columnwidth]{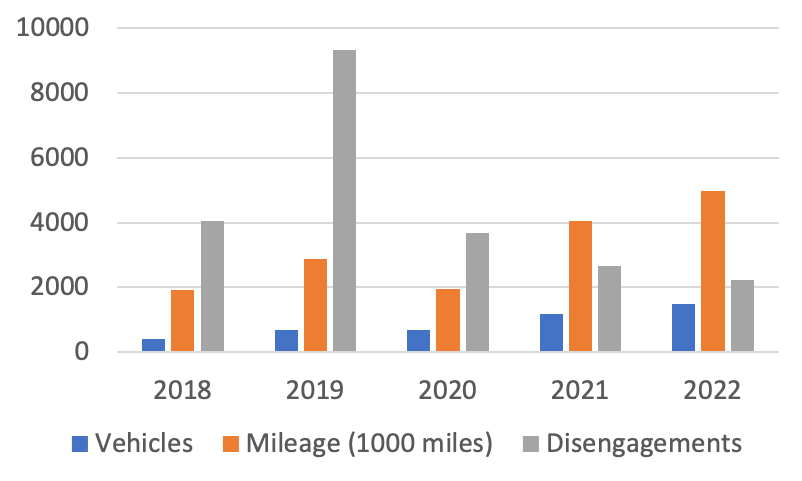}
\caption{California: the number of self-driving vehicles, mileage and disengagements from 2018 to 2022~\cite{disengagement}.}
\vspace{10pt}
\label{fig:disen}
\end{figure}

Figure ~\ref{fig:disen} illustrates Autonomy 1.0's scalability problems using autonomous driving operation data from California from 2018 to 2022. Over the past five years, although enormous amount of investment has been poured into autonomous driving, we did not observe significant growth of the number of vehicles under operation, which increased only from 400 in 2018 to 1,500 in 2022. The operation mileage per year increased only from 2 million miles to 5 million miles. Most importantly, there are still over 2,000 disengagement incidents per year. Given this trend in Autonomy 1.0, we are still years away from serious commercial operations of autonomous vehicles.
 
Autonomy 1.0 is modular and consists of functional modules such as sensing, perception, localization, high-definition maps, prediction, planning and control~\cite{yu2020building}, each further consists of several functional sub-modules integrated by explicit and hand-crafted logic. Most decision-making tasks, such as planning, which is responsible for generating optimal and drivable paths, are solved with constraint optimization under a set of hand-tuned rules. When a disengagement incident happens, engineers usually have to go through a long process of debugging to identify which specific module or rule may have been the root cause of the disengagement, then optimize that module or develop logic changes to handle the specific problem. Often, due to intricate dependency and coupling among modules or rules, the new software version leads to other problems that need to be addressed, thus greatly slowing down development process. The Autonomy 1.0 software stack over time became a complicated collection of ad-hoc rules and a set of interdependent modules for handling various long-tail events, which has been increasingly difficult to debug, maintain and evolve for improved performance.

Taking the open-source project Apollo~\cite{apollo} as an example, its perception module alone consists of multiple individual leaning-based sub-modules to accomplish object detection in 2D images, LiDAR point cloud segmentation, traffic light detection, lane detection, and others. To integrate information from these perception sub-modules, a post-processing module then fuses 2D and 3D information and outputs an integrated representation of the environment to the downstream prediction module. The planning module makes decisions and plans routes based on the data from the prediction, localization, and map modules. These modules often have strong dependencies among themselves. Making changes to one module not only impacts the overall system performance, possibly violating real-time constraints and resource allocation, but also impacts the algorithmic performance of other downstream modules due distributional shift of data.
The whole system has become complicated and even brittle, demanding enormous amount of engineering resources to maintain, let alone to scale.

We summarize the three Autonomy 1.0's major scalability bottlenecks below.

\begin{itemize}
    \item \textbf{\textit{Complexity Bottleneck}}: The design of autonomy 1.0 systems demands extensive engineering efforts to define software interfaces, distribute data among modules, and map various workloads in a heterogeneous computing system. It is challenging, given the complexity, to debug and continuously update the software stack.
    The myriad of components also make it challenging to schedule tasks and optimize the latency of the unwieldy stack at run-time.
    As a result, typical autonomy 1.0 systems exhibit large latency variations~\cite{yu2020building}, which can harm the reliability of the autonomous driving system.
    \item  \textbf{\textit{Human-Data Bottleneck}}: Autonomy 1.0 systems depend on fleets of physical vehicles operated by humans to collect data and perform system-level tests. This is a time-consuming and expensive process that is difficult to scale out.
    The scalability issue will only get worse as increasingly more modules of autonomy stack adopt data-driven approaches, which requires \textit{continuous} collection and labeling, because any specific instance of the recorded data reflects only a particular subset of the world states.
    \item \textbf{\textit{Generalization Bottleneck}}: Autonomy 1.0 systems consist rule-based processing logic and hand-crafted interfaces, which makes them difficult to generalize to new environments. This is because the complexity and diversity of real-world environments makes it difficult to design the autonomy system to anticipate all possible challenging scenarios, whether for perception or planning. As a result, autonomy 1.0 systems are often over-fitted to frequently operated regions and common situations. To handle new environments and newly encountered rare cases, additional changes to the system are required, which is increasingly difficult and time-consuming.
\end{itemize}

\section{Autonomy 2.0: Scalability is Everything}
\label{sec:a2}

Recent research breakthroughs in artificial intelligence, such as Transformer \cite{vaswani2017attention}, large language models (LLM) \cite{brown2020language} and offline reinforcement learning \cite{levine2020offline}, have sparked new ideas in architecture design, data and model infrastructure, and engineering practices of autonomous driving, leading to a new development paradigm, which we dub Autonomy 2.0.

\begin{figure*}
\centering
\includegraphics[width=1\linewidth]{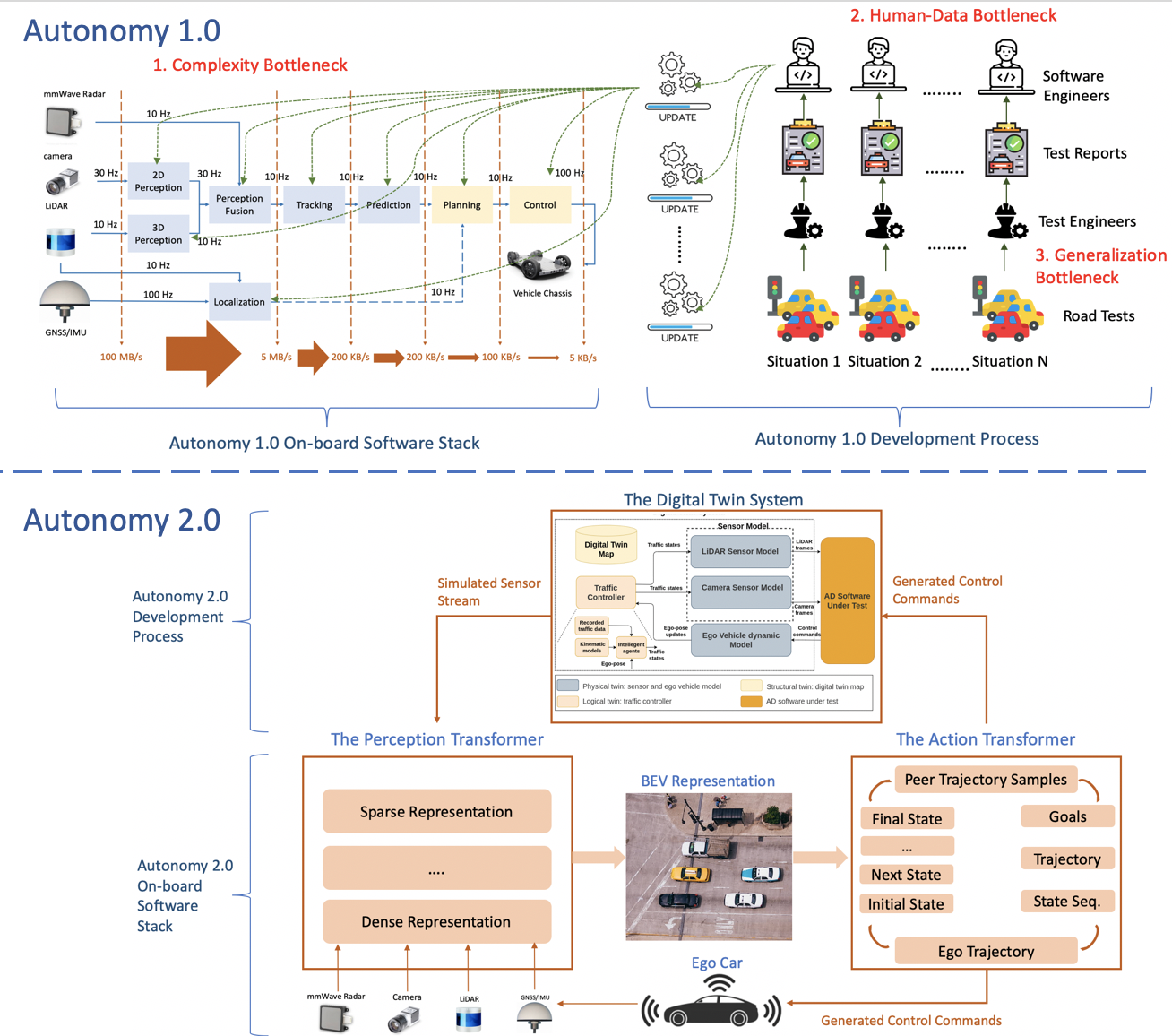}
\caption{Autonomy 1.0 vs. 2.0 Architectures.}
\vspace{10pt}
\label{fig:arch}
\end{figure*}

The key of Autonomy 2.0 is scalability, which is delivered through two ingredients: 1) a software stack that improves continuously with increasing scale of data and compute resources. 2) a simulation paradigm based on digital twins for algorithmic exploration using large-scale, real-time, realistic data before deployment. Figure ~\ref{fig:arch} illustrates the differences between Autonomy 1.0 and Autonomy 2.0 system architectures. Table ~\ref{table:a2} summarizes how Autonomy 2.0 addresses the three bottlenecks in Autonomy 1.0. 

\vspace{5pt}
\begin{table*}[htbp]
\caption{Summary of Autonomy 1.0 vs. Autonomy 2.0}
\begin{center}
\begin{tabular}{ccc}
 \hline & \textbf{Autonomy 1.0} & \textbf{Autonomy 2.0} \\ 
 \hline\hline
 \textbf{Complexity Bottleneck} & \begin{tabular}{@{}c@{}}numerous functional modules \\ heavy dependencies among modules \\ complex real-time constraints\end{tabular} & \begin{tabular}{@{}c@{}}two transformer modules \\ well-defined software interfaces \\ simple real-time constraints\end{tabular} \\
 \hline
 \textbf{Human-Data Bottleneck} & \begin{tabular}{@{}c@{}}data generation through road tests \\ scale with human labor\end{tabular} &  \begin{tabular}{@{}c@{}}data generation through digital twin \\  scale with computing resources \end{tabular} \\ 
 \hline
 \textbf{Generalization Bottleneck} & \begin{tabular}{@{}c@{}}task-specific hand-crafted functions \\ complicated software stack\end{tabular} & \begin{tabular}{@{}c@{}}task-agnostic learning-based updates \\ simple and stable software stack\end{tabular} \\
 \hline
\end{tabular}
\label{table:a2}
\end{center}
\end{table*}

\subsection{Learning-Native Software Stack}

Any autonomous machine performs two main tasks: perception and action, reflecting the natural dichotomy of the past and the future. The perception task observes the environment and infers its current state based on observations so far. The action task, based on these observations, chooses an appropriate sequence of actions to achieve goals while considering how the environment may evolve in the near future.

The software stack in Autonomy 2.0, thus, naturally consists of a perception module and an action module. Unlike in Autonomy 1.0 where each module is implemented by a number of sub-modules, there is a strong evidence that the two modules, in Autonomy 2.0, will each be implemented as a single large deep learning model, likely based on transformer or its variants due to their ability to generalize, as demonstrated in their recent successes in LLMs.

\paragraph{Benefits.} 
Before describing how the two-model architecture will look like in Autonomy 2.0, we will first discuss why such an architectural design choice is key to scalability.


The two-model architecture addresses the Complexity Bottleneck by drastically reducing the amount of code that needs to be maintained and reasoned about.
Figure \ref{fig:ad12}a) compares the lines of code in the Apollo Perception module~\cite{apollo}, which represents the Autonomy 1.0 approach, with an example of the perception module in Autonomy 2.0, BEVFormer ~\cite{li2022bevformer}. The Apollo Perception module's size is ten times larger than BEVFormer, and BEVFormer has achieved state of the art perception results.

The software architecture also handles corner cases through data-driven model learning instead of hand-crafted logic, and thus address the Generalization Bottleneck in Autonomy 1.0.
In Figure \ref{fig:ad12}b), we analyze over 400 issues associated with the Apollo planning modules, 47\% of the issues are related to Apollo failing to handle a specific usage case, and 30\% of the issues are related to software engineering problems such as interfaces with other modules. In Autonomy 1.0, many hand crafted rules are implemented to handle specific use cases. As the rules accumulate, software quality naturally becomes an issue.

\paragraph{Architectural Design.}
The perception and action modules have different goals and traditionally require distinctive algorithmic approaches.
The perception module is trained using supervised learning and self-supervised learning to infer one unique ground truth of world states.
In contrast, the action module needs to search and choose from many acceptable action sequences, while anticipating the behaviors of other agents.
Therefore, the action module makes use of methods from reinforcement learning, imitation learning, and model predictive control.

Interestingly, while the fundamental distinctions of the two modules have not changed in Autonomy 2.0, there is a growing convergence of the implementation of the two modules: recent successes of large language models (LLM) \cite{brown2020language} to comprehend a large amount of information to perform multiple sub-tasks suggest that both modules can be implemented using a similar architecture based on Transformer \cite{vaswani2017attention}.

Transformer is a great algorithmic substrate for both the perception and action modules because of its ability to generalize.
For perception, a transformer can effectively fuse perceptual data from multiple sensors and multiple moments into a unified representation, avoiding information loss from sparsification and module serialization. For action, the sequential nature of transformer makes it a perfect fit for processing and generating temporal data, especially for sampling multiple possible future paths.

\underline{Perception.}
In Autonomy 1.0, the perception module consists of multiple DNNs, each trained separately to support individual tasks such as 2D/3D object detection, segmentation, and tracking.
In contrast, the perception module in Autonomy 2.0 uses a single transformer backbone to provide a unified representation of the ego-vehicle's environment (e.g., 2D Bird's Eye View (BEV) \cite{li2022bevformer, li2022hdmapnet, li2022bevdepth} or 3D occupancy \cite{wei2023surroundocc, wang2023openoccupancy, huang2023tri}), which is then attached to a number of decoder ``heads'', each of which is tuned for an individual task.

This single-transformer approach toward the perception module has been gaining popularity across the AV industry. 
For instance, this is the approach described by Tesla engineers in their ``AI Day 2022'' event ~\cite{tesla}, and has been deployed by another leading intelligent electric vehicle company XPENG ~\cite{xpeng}.


\underline{Action.} The action module anticipates a combinatorially large number of possible ``world trajectories'', hypothesizes multiple action sequences, and evaluates them to send the optimal one to actuators.
In Autonomy 1.0, the action module is implemented as a set of sub-modules for prediction, planning, and control.
The action module in Autonomy 2.0 is end-to-end learned using transformer-inspired architectures for sequential decision making \cite{chen2021decision, janner2021offline, chen2022transdreamer, gupta2022metamorph, chen2022learning}.

The action transformer incorporates two models: a policy model and a world model.
First, the pre-trained, transformer-based \textit{policy model} leverages the large amount of historical data for agent behavior prediction and ego vehicle decision making and trajectory planning \cite{yu2022leverage, herzog2023deep, ball2023efficient}. 
Second, the world model is essentially a behaviorally realistic simulator (validated against real-world data) of the world.
The two models are connected with a closed-loop in the transformer so that the policies can be fine-tuned online \cite{nakamoto2023cal}.



\subsection{Digital-Twin Based Development and Deployment}

Autonomy 1.0 relies almost exclusively on human efforts for tasks such as manual data labeling and physical testing, posing a scalability bottleneck.
Autonomy 2.0 addresses the ``Human-Data Bottleneck'' using an emerging simulation technology called digital twins, where a virtual representation acts as the counterpart of the physical world.
As highlighted by the recent National Artificial Intelligence R\&D Strategic Plan 2023 published by the White House~\cite{aistrategy2023}, digital twins have fueled many real-world applications (e.g., urban planning/management of smart cities and additive manufacturing), and is a main strategy to sustain AI technologies.

Under the digital-twins paradigm, one instruments the physical system to collect real-world, real-time data, which is then interactively shared with the digital counterpart.
In the digital world, one could further synthesize scenarios (e.g., traffics) with a statistically significant fidelity with a similar behavioral distributions as that in human driving behaviors.

\begin{figure*}
\centering
\includegraphics[width=0.75\linewidth]{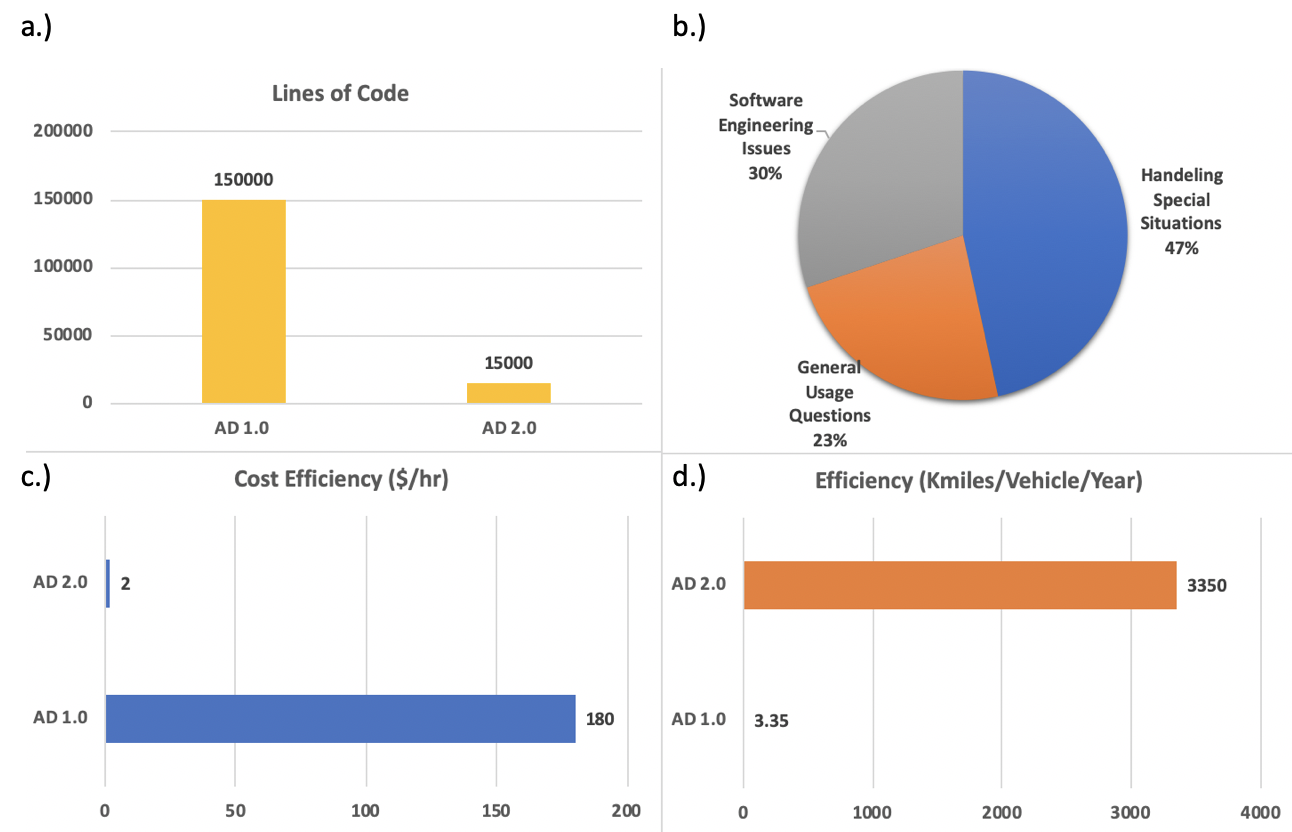}
\caption{Comparison of Autonomy 1.0 vs. Autonomy 2.0: a.) lines of code in Apollo Perception module (Autonomy 1.0) vs. in BEVFormer (Autonomy 2.0); b.) breakdown of issues in Apollo Planning module; c.) R\&D cost efficiency comparisons d.) R\&D efficiency comparisons}
\vspace{10pt}
\label{fig:ad12}
\end{figure*}

Developing and testing autonomous driving software using synthesized virtual scenarios accelerates the evaluation process by $10^3$ to $10^5$ times ~\cite{feng2023dense} and reduces the testing costs by two orders of magnitude ~\cite{yu2022autonomous} compared to the physical-only approach in Autonomy 1.0.
Figure \ref{fig:ad12}c) demonstrates the R\&D cost efficiency in Autonomy 1.0, which costs \$180/hr through physical testing, vs. in Autonomy 2.0, which costs \$2/hr through virtual testing, an 100-fold improvement~\cite{yu2022autonomous}. Figure \ref{fig:ad12}d) demonstrates the R\&D efficiency in Autonomy 1.0, which takes around 3 kilo miles per physical vehicle per year through physical testing\cite{disengagement}, vs. in Autonomy 2.0, which takes over 3 million miles per virtual vehicle per year through simulation, a 1000-fold improvement~\cite{feng2023dense}. Combining these two factors would bring over $10^5$ times improvement under the same engineering investment in Autonomy 2.0, and scalability is thus only constrained by the available compute resources instead of number of engineers, effectively eliminating the human-data bottleneck. 

\section{Summary}
\label{sec:sum}

The autonomy economy, or the use of autonomous machines to provide goods and services, will fuel the world's economic growth in the coming decades.
Huge investments are pouring into the autonomy economy. Such a huge investment will only be justified if autonomous machines can reach, and provide utility for, every person on planet.
Similar to today's digital economy, scalability will necessarily be the winning formula in this process.
The current practice of developing and deploying autonomous machines carries the historical baggage of complexity bottleneck, human-data bottleneck, and generalization bottleneck, and is thus unscalable.
We must start from a clean slate and rethink the architecture design of autonomous machines.


We posit that Autonomoy 2.0 will embrace a learning-native software stack, which addresses the complexity bottleneck through software simplicity and addresses the generalization bottleneck through end-to-end learning.
The digital twins technologies will have to be integrated throughout the development, evaluation, and deploymemt cycle in Autonomy 2.0 to address the human-data bottleneck.


\bibliographystyle{ieeetr}
\footnotesize{%
\bibliography{ref}
}
\end{document}